\pdfoutput=1

\documentclass[11pt]{article}

\usepackage{EMNLP2022}

\usepackage{times}
\usepackage{latexsym}

\usepackage[T1]{fontenc}

\usepackage[utf8]{inputenc}

\usepackage{microtype}

\usepackage{inconsolata}

\usepackage{xspace}
\usepackage{graphicx}
\usepackage{algorithmic}
\usepackage{textcomp}
\usepackage{xcolor}
\usepackage{booktabs}
\usepackage{multirow}
\usepackage{amsmath}
\usepackage{xparse}
\usepackage{colortbl}
\usepackage{array}
\usepackage{stackengine}
\usepackage{makecell}

\newcommand{\pasta}{\textsc{Pasta}\xspace}
\newcommand\xrowht[2][0]{\addstackgap[.5\dimexpr#2\relax]{\vphantom{#1}}}
\newcommand{\eg}{{e.g.,}\xspace}
\newcommand{\ie}{{i.e.,}\xspace}
\newcommand{\aka}{{a.k.a.}\xspace}

\newcommand{\stitle}[1]{\vspace{1mm}\noindent{\bf #1}}

\newcommand{\term}[1]{${\tt #1}$\xspace}

\newcommand{\sstab}{\vspace{0.2ex}\noindent}

\NewDocumentCommand{\preslav}{ mO{} }{\textcolor{blue}{\textsuperscript{\textit{Preslav}}\textsf{\textbf{\small[#1]}}}}

\NewDocumentCommand{\nan}{ mO{} }{\textcolor{violet}{\textsuperscript{\textit{Nan}}\textsf{\textbf{\small[#1]}}}}

\title{\pasta: Table-Operations Aware Fact Verification\\ via Sentence-Table Cloze Pre-training}

\author{Zihui Gu$^1$, Ju Fan$^1$, Nan Tang$^2$, Preslav Nakov$^3$, Xiaoman Zhao$^1$\thanks{~~Xiaoman Zhao is the corresponding author.}~, Xiaoyong Du$^1$ \\
  $^1$Renmin University of China, Beijing, China, \\
  $^2$Qatar Computing Research Institute, HBKU, Doha, Qatar \\
  $^3$Mohamed bin Zayed University of Artificial Intelligence, Abu Dhabi, UAE \\
  \texttt{$^1$\{guzh,fanj,xiaomanzhao,duyong\}@ruc.edu.cn}, 
  \\ \texttt{$^2$ntang@hbku.edu.qa}, \texttt{$^3$preslav.nakov@mbzuai.ac.ae}\\}

\begin{document}
\maketitle

\begin{abstract}
Fact verification has attracted a lot of research attention recently, \eg in journalism, marketing, and policymaking, 
as misinformation and disinformation online can sway one's opinion and affect one's actions. 
While fact-checking is a hard task in general, in many cases, false statements can be easily debunked based on analytics over tables with reliable information.
Hence, table-based fact verification has recently emerged as an important and growing research area.
Yet, progress has been limited due to the lack of datasets that can be used to pre-train language models (LMs) to be aware of common table operations, such as aggregating a column or comparing tuples. 
%
To bridge this gap, in this paper we introduce \pasta, a novel state-of-the-art framework for table-based fact verification via pre-training with synthesized sentence--table cloze questions.
%
In particular, we design six types of common sentence--table cloze tasks, including \term{Filter}, \term{Aggregation}, \term{Superlative}, \term{Comparative}, \term{Ordinal}, and \term{Unique}, 
based on which we synthesize a large corpus consisting of 1.2 million sentence--table pairs from WikiTables.
\pasta uses a recent pre-trained LM, DeBERTaV3, and further pre-trains it on our corpus.
Our experimental results show that \pasta achieves new state-of-the-art performance on two table-based fact verification benchmarks: TabFact and SEM-TAB-FACTS. In particular, on the complex set of TabFact, which contains multiple operations, \pasta largely outperforms the previous state of the art by {\bf 4.7} points (85.6\% vs. 80.9\%), and the gap between \pasta and human performance on the small TabFact test set is narrowed to just {\bf 1.5} points (90.6\% vs. 92.1\%).
\footnote{The pre-trained model, the pre-training corpus, and the source code are released at\\ \url{https://github.com/ruc-datalab/PASTA}}
%
\end{abstract}


\section{Introduction}

Fact verification, which checks the factuality of a statement, is crucial for journalism~\cite{fakenews}, and is increasingly being  applied in other fields~\cite{application1,application2}.
According to Duke Reporters' Lab, there are $300+$ active certified fact-checking organizations worldwide.\footnote{\url{http://reporterslab.org/fact-checking-count-tops-300-for-the-first-time/}}
%
%
%

Automatic and explainable approaches, \aka reference-based approaches, are widely used to assist fact-checkers. They verify the input statement against a trusted source, such as relevant passages from Wikipedia~\cite{DBLP:conf/www/PopatMSW17,DBLP:conf/naacl/ThorneVCM18,DBLP:conf/acl/ShaarBMN20}.
Recently, table-based fact verification has been extensively studied~\cite{tabfact,logicfactchecker,tapasfv} due to the wide availability of tabular data.
%
%

\begin{figure}[tbh]
	\centering
	\includegraphics[width=0.98\columnwidth]{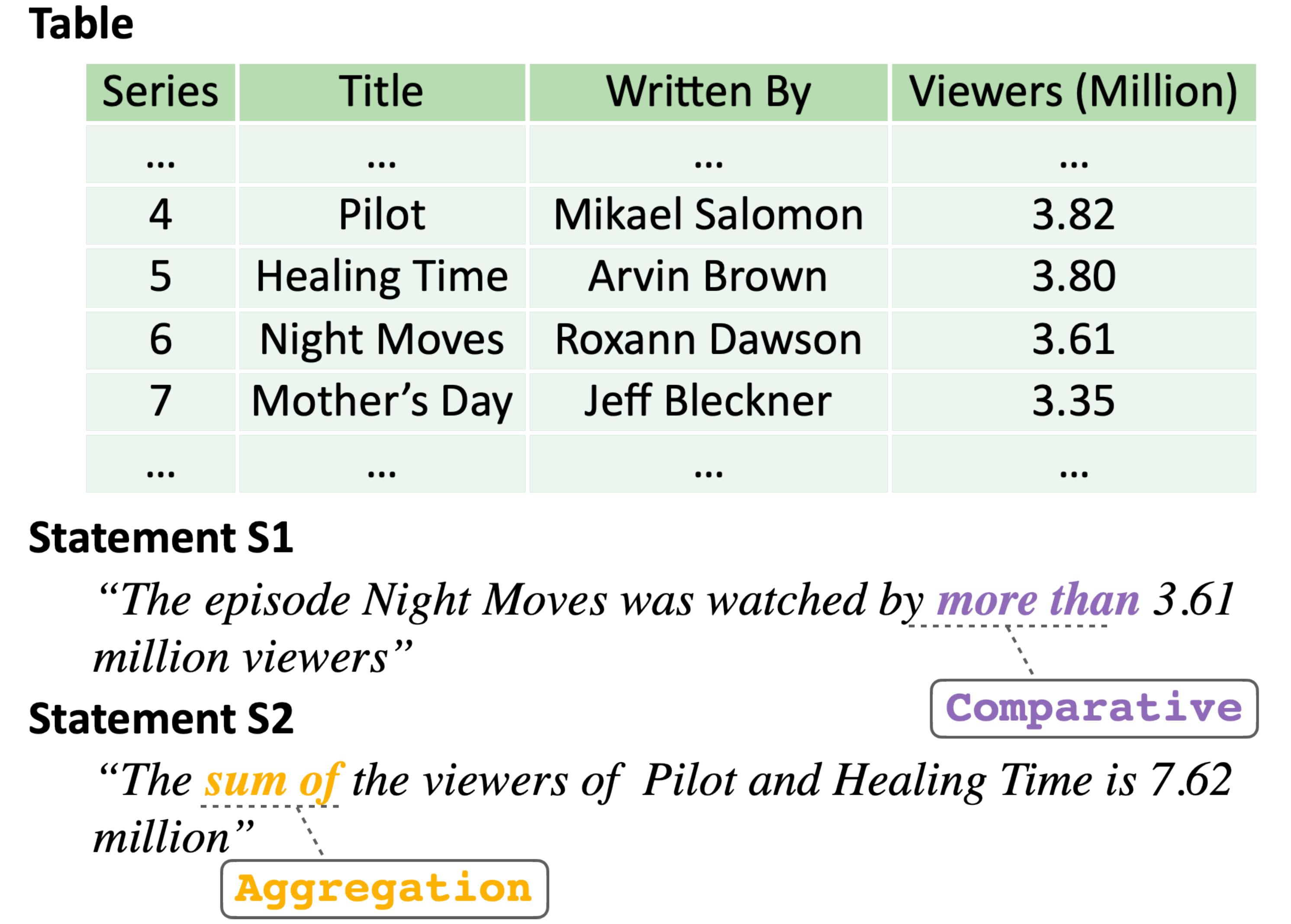}
	\caption{An example of table-based fact verification.}
	\label{fig:fv_example}
\end{figure}

Evidently, performing fact verification over tables requires the ability to reason about \textbf{table-based operations}, such as aggregating the values in a column or comparing tuples.
%
%
For example, for the statement $S_1$ in Figure~\ref{fig:fv_example}, it is desirable to reason about the {operation} over table $T$ that compares the viewers of \term{Night~Moves} with {3.61} to determine whether $S_1$ is \emph{entailed} or \emph{refuted} by $T$.

Most previous work~\cite{tapas,salience,joint} leverages pre-trained language models (LMs)~\cite{bert,roberta}, which are originally designed for unstructured data, 
and have a key limitation of overlooking such operations.
Some approaches~\cite{logicfactchecker,progvgat} attempt to explicitly capture the operations by generating a logical form (\eg a tree) containing the operations from the statement via semantic parsing techniques.
However, such approaches face the problem of ``spurious programs''~\cite{tabfact}, due to weak supervision signals in semantic parsing.


To address the above issues, we propose \pasta,
a table-operations aware approach. Instead of relying on semantic parsing, \pasta captures table-based operations by designing a novel \textbf{sentence--table cloze pre-training} strategy that better guides LMs to reason about table-based operations. 

We tackle two challenges 
for pre-training LMs towards supporting table-based fact verification:
\begin{itemize}
    \item {\bf Challenge 1:} What types of tasks should be designed, so as to pre-train (or teach) LMs to be aware of operations over tables? 
    \item {\bf Challenge 2:} How to obtain a large-scale and high-quality corpus for pre-training?
\end{itemize}

To address {\bf Challenge 1}, \pasta automatically synthesizes sentence--table cloze questions. It first synthesizes operations from tables, and then generates cloze tasks by masking the key tokens corresponding to the table-based operations, \eg ``\emph{more than}'' and ``\emph{sum of}'' in Figure~\ref{fig:fv_example}. Then, LMs are pre-trained to predict the masked operation-aware tokens based on the tables.

Regarding {\bf Challenge 2}, \pasta uses a large table collection, WikiTables~\cite{wikitables}, and for each table, it synthesizes a diverse set of cloze tasks with six types of table-based operations, including \term{Filter}, \term{Aggregation}, \term{Superlative}, \term{Comparative}, \term{Ordinal}, and \term{Unique}. 

For implementation, \pasta uses a recent pre-trained LM, DeBERTaV3~\cite{debertav3,deberta} with better positional encoding of the input.
To cope with the limited input length of DeBERTaV3, we introduce a {select-then-rank} strategy for large tables, which further improves the performance.
%




In sum, we make the following contributions.

\begin{itemize}
    \item We propose \pasta, a table-operations aware fact verification approach without explicitly generating logical forms.
    \item We propose a new benchmark for pre-training LMs to be aware of common table-based operations by automatically synthesizing sentence-table cloze questions from WikiTables. In particular, we synthesize a large corpus consisting of 1.2 million sentence-table cloze questions, which we release for future research.
    \item We evaluate \pasta, which is DeBERTaV3 pre-trained with our table-operations aware pre-training approach, on two widely-adopted table-based fact verification benchmark datasets, TabFact~\cite{tabfact} and SEM-TAB-FACTS~\cite{semtabfact}. The experimental results show that \pasta achieves new state-of-the-art (SOTA) results on the two datasets. In particular, on the complex set of TabFact that contains multiple operations, \pasta outperforms the previous SOTA by {\bf 4.7} points (85.6\% vs. 80.9\%), and the gap between \pasta and human performance on the small test set is narrowed to {\bf 1.5} points (90.6\% vs. 92.1\%).
\end{itemize}

\section{Preliminaries}


\subsection{Problem Formulation}
%
Let $T$ be a table with $m$ columns and $n$ rows. 
Let $T_{i,j}$ denote the cell in the $i$-th column and $j$-the row of $T$. 
Let $S$ be a natural language (\term{NL}) statement. 

The problem of \textbf{table-based fact verification} is formulated as follows: Given an \term{NL} statement $S$ and a table $T$, it determines whether statement $S$ can be \emph{entailed} or \emph{refuted} by table $T$\footnote{This formulation is in line with TabFact~\cite{tabfact}. Note that other variants of this problem are also possible, \eg with a third option {\em not sure}.}.

See Figure~\ref{fig:fv_example} for an example table about movies and their viewers, and two statements where $S_1$ contains a \term{Comparative} operation and $S_2$ contains an \term{Aggregation} operation.



\subsection{DeBERTa for Sentence-Table Encoding}
\label{sec:deberta}

Inspired by the success of BERT-like models~\cite{bert,roberta,electra} in natural language understanding (NLU) tasks, many existing studies leverage pre-trained LMs for table understanding, achieving superior results~\cite{tabsearch,joint}.
In this paper, we apply DeBERTa~\cite{deberta} for sentence-table encoding, as it can effectively capture positional information of the input with its positional encoding scheme, which is useful for sentence-table encoding.

Given an input token at position $i$, DeBERTa represents it using two vectors, $\{H_i\}$ and $\{P_{i|j}\}$, to represent its content and relative position with respect to the token at position $j$. For a single-head self-attention layer, DeBERTa~\cite{deberta} represents the disentangled self-attention mechanism as follows:
\begin{equation} \nonumber
	\begin{aligned}
	&Q_c=HW_{q,c}, K_c=HW_{k,c}, V_c=HW_{v,c}  \\ 
	&Q_r=PW_{q,r}, K_r=PW_{k,r} \\
	&\Tilde{A}_{i,j}=Q_i^c{K_j^c}^\mathrm{T}+Q_i^c{K_{\delta(i,j)}^r}^\mathrm{T}+K_j^c{Q_{\delta(j,i)}^r}^\mathrm{T},
	\end{aligned}
\end{equation}
%
%
where $\Tilde{A}_{i,j}$ represents the attention score between token $i$ and token $j$.
The content vector $H$ is projected by the matrices $W_{q,c}, W_{k,c}, W_{v,c}\in R^{d\times d}$ to generate the projected content vectors $Q_c$, $K_c$ and $V_c$, respectively,
$P\in R^{2k\times d}$ is the relative position embedding vector, and $\delta(i,j)$ is the relative distance from token $i$ to token $j$.
Similarly to $H$, $P$ is projected by the matrices $W_{q,r}, W_{k,r}\in R^{d\times d}$ to generate the projected position vectors $Q_r$ and $K_r$, respectively. 

In our implementation, we adopt the latest version DeBERTaV3~\cite{debertav3}, which improves DeBERTa by further pre-training with replaced token detection (RTD) to jointly encode a sentence and a table. Unlike \term{NL}, tables have distinct structural information that is difficult to capture by a pre-trained LMs. Therefore, we use special symbols to inject the structural information into an \term{NL} sentence. Specifically, we linearize the table $T=\{T_{i,j}|i\leq m, j\leq n\}$ into sentence $S_T$ = \term{[Header]} $T_{0,0}$ | $T_{0,1}$ … | $T_{0,n}$ \term{[Row]} $T_{1,0}$ | $T_{1,1}$ … \term{[Row]} $T_{i,0}$ | $T_{i,1}$ … | $T_{m,n}$. We use \term{[Header]} to indicate the beginning of the headers and \term{[Row]} to indicate the beginning of each row. Inspired by~\citealp{tapex}, we also use ``\term{|}'' to separate each cell. Afterwards, we concatenate the statement $S$ and the linearized table $S_T$. 

\section{Our \pasta Model}

\begin{figure*}[t!]
	\centering
	\includegraphics[width=1\textwidth]{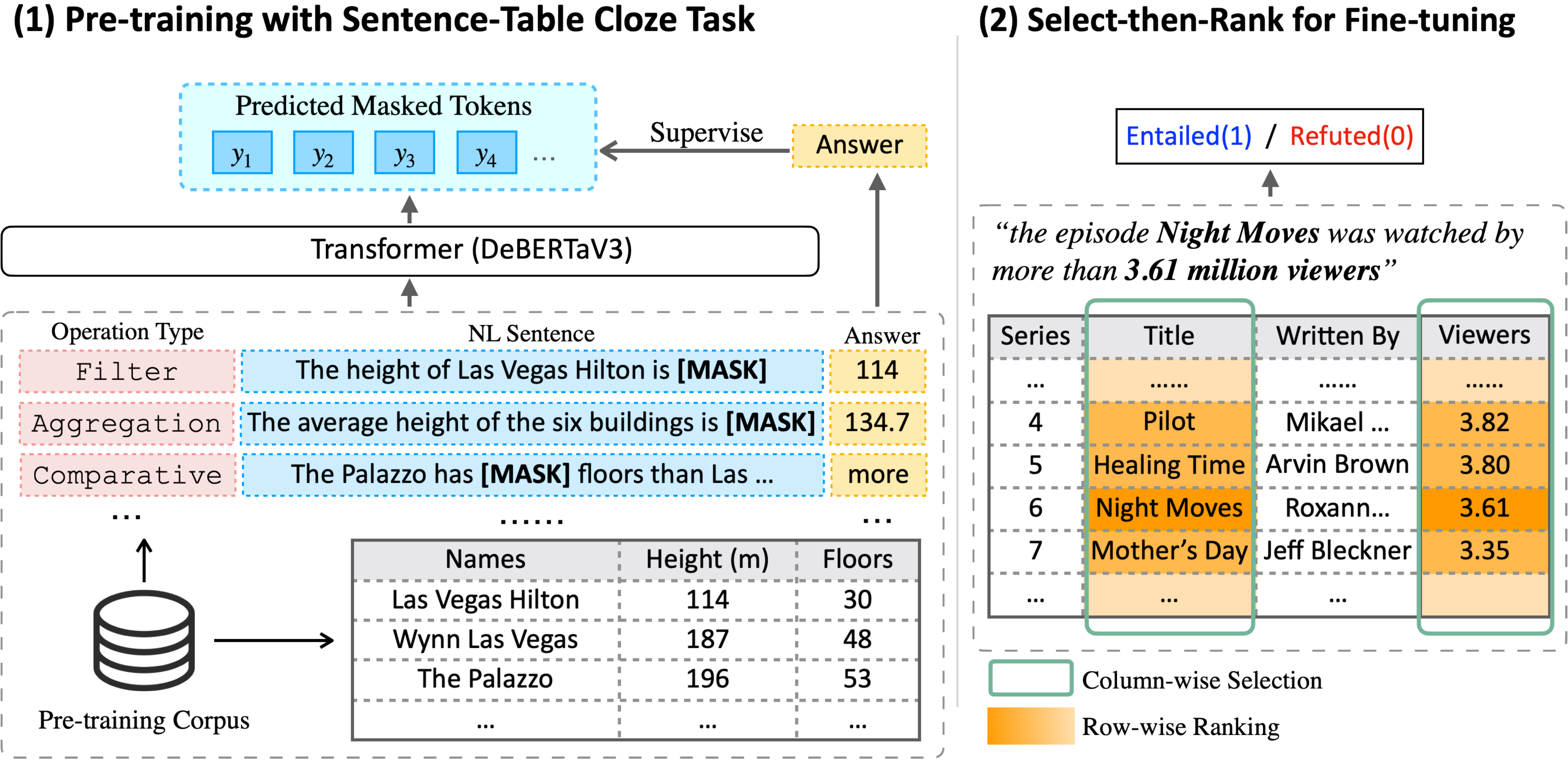}
	\caption{An overview of the pre-training and fine-tuning procedures of \pasta.}
	\label{fig:overview}
\end{figure*}

Figure~\ref{fig:overview} gives an overview of our \pasta framework, which follows the pre-training--fine-tuning framework. 
For {\bf pre-training}, we guide our model to understand sentences and to perform table-based operations
(\eg \term{Aggregation}) to complete \textbf{\textit{synthesized cloze tasks}} in the sentence. For {\bf fine-tuning}, we apply a \textbf{\textit{select-then-rank}} strategy to trade off between sizes of large tables and the limited input length of DeBERTaV3.
Next, we will first present our sentence-table cloze task and pre-training corpus generation in Sections~\ref{sec: pretrain_task} and \ref{sec:pretrain_corpus}, respectively. We will then discuss our fine-tuning strategy in Section~\ref{sec:finetune}.

%

\subsection{Sentence-Table Cloze Pre-training}
\label{sec: pretrain_task}

An essential ability for our fact verification model is to be capable of reasoning about \textbf{table-based operations} on the entities of the table.
%
Although it is difficult to enumerate all the expressions of real-world statements, the types of operations (\eg \term{Unique}) expressed by statements could be limited. Therefore, it is possible and reasonable for the model to understand these operations. To this end, we propose a table-operations aware pre-training task which can guide the model to understand table-aware operations expressed by the statements.

Inspired by the Masked Language Modeling (MLM)~\cite{bert}, we design a \textbf{cloze task} to pre-train the model's ability to reason about operations over tables.
However, the key difference is that we do not use the \emph{random masking} strategy in MLM due to the following reasons.
First, masking a specific cell of the table and training the model to predict it is difficult, because, unlike the words in a sentence, the contents of a cell may not be predictable from the contents of the surrounding cells. Moreover, the content of an individual cell may be useless for determining whether statement $S$ can be entailed or refuted by table $T$.
Second, not every token in the sentence needs to be predicted. For example, in the sentence \textit{``The Palazzo has more floors than Las Vegas Hilton.''}, tokens like \textit{has} and \textit{the}, which can be easily predicted from contextual information, are not worth learning for the model because this kind of ability is already captured by pre-trained LMs.



To pre-train the model to be aware of table operations, we propose to \textbf{mask operation-aware tokens} in the sentence, which
need to meet two requirements: (\emph{i})~to appear in the sentence and to correspond to table-based operations, and (\emph{ii})~to be predictable by reasoning over tables. For example, in the above sentence, as the model needs to find the numbers of floors for ``\textit{Las Vegas Hilton}'' and for ``\textit{The Palazzo}'' in the table and then to compare them, ``\textit{more}'' is the operation-aware token that the model needs to predict.


To cover common types of operations in pre-training, we refer to the operations list defined in LPA~\cite{tabfact}. We design \textbf{six sentence-table cloze tasks} according to various operation types: \term{Filter}, \term{Aggregation}, \term{Superlative}, \term{Comparative}, \term{Ordinal}, and \term{Unique}. Figure~\ref{fig:overview} shows some example cloze tasks of operation types, \term{Filter}, \term{Aggregation} and \term{Comparative}. The answer to a cloze task, \ie operation-aware tokens to be predicted, may be a specific table cell (\eg ``\emph{114}'') or the result of a series of calculations (\eg ``\emph{134.7}'', ``\emph{more}'').
Note that we assume that only atomic operation types need to be learned in pre-training, and various combinations and expressions are left to fine-tuning. 
Thus, only one type of operation-aware token is masked in each statement.

We formally define the table-operations aware pre-training task as follows: Given a sentence $S=\{x_i\}$ and a table $T=\{T_{i,j}|i\leq m,j\leq n\}$, we corrupt $S$ into $\Tilde{S}$ by masking the operation-aware span of tokens $S_{span}=\{\Tilde{x_i}\}\subset S$ in it, and then we train an LM parameterized by $\theta$ to reconstruct $S$ by predicting the masked tokens $\{\Tilde{x_i}\}$, \ie optimizing the following objective:
\begin{align*}
L_{\pasta} &= -\log \mathop{p_\theta}(S|\Tilde{S},T) \\
&= -\sum^{}_{\Tilde{x_i}\in S_{span}}\log\mathop{p_\theta}(\Tilde{x_i} = x_i|\Tilde{S}, T) 
\end{align*}

\subsection{Pre-training Corpus Generation}
\label{sec:pretrain_corpus}

Next, we introduce our strategy for generating the pre-training corpus consisting of sentence--table pairs. According to the table-operations aware pre-training task described in Section~\ref{sec: pretrain_task}, the difficulty of corpus generation is how to collect a large scale of sentence--table pairs and how to identify the operation-aware tokens in each sentence. To solve these problems, we propose an automatic data generation method, which consists of table collection and sentence generation. In addition, we also introduce a probing-based sentence polishing method to make it more fluent and natural.

\stitle{Table Collection.} Inspired by previous work~\cite{tapas,joint}, we use WikiTables,\footnote{\url{http://websail-fe.cs.northwestern.edu/TabEL/}} which contains Web tables extracted from Wikipedia. Concretely, we only select well-formed relational tables 
%
that contain headers and at least one numeric column that can be used for operations. Moreover, considering the maximum input length (512 tokens) of our pre-trained LM, we filter out all tables with more than 500 cells.
%
Based on the above process, we obtain a total of 580K tables from WikiTables, and we then randomly select 20K tables to improve the efficiency of pre-training.

\begin{figure}[t!]
	\centering
	\includegraphics[width=1.\columnwidth]{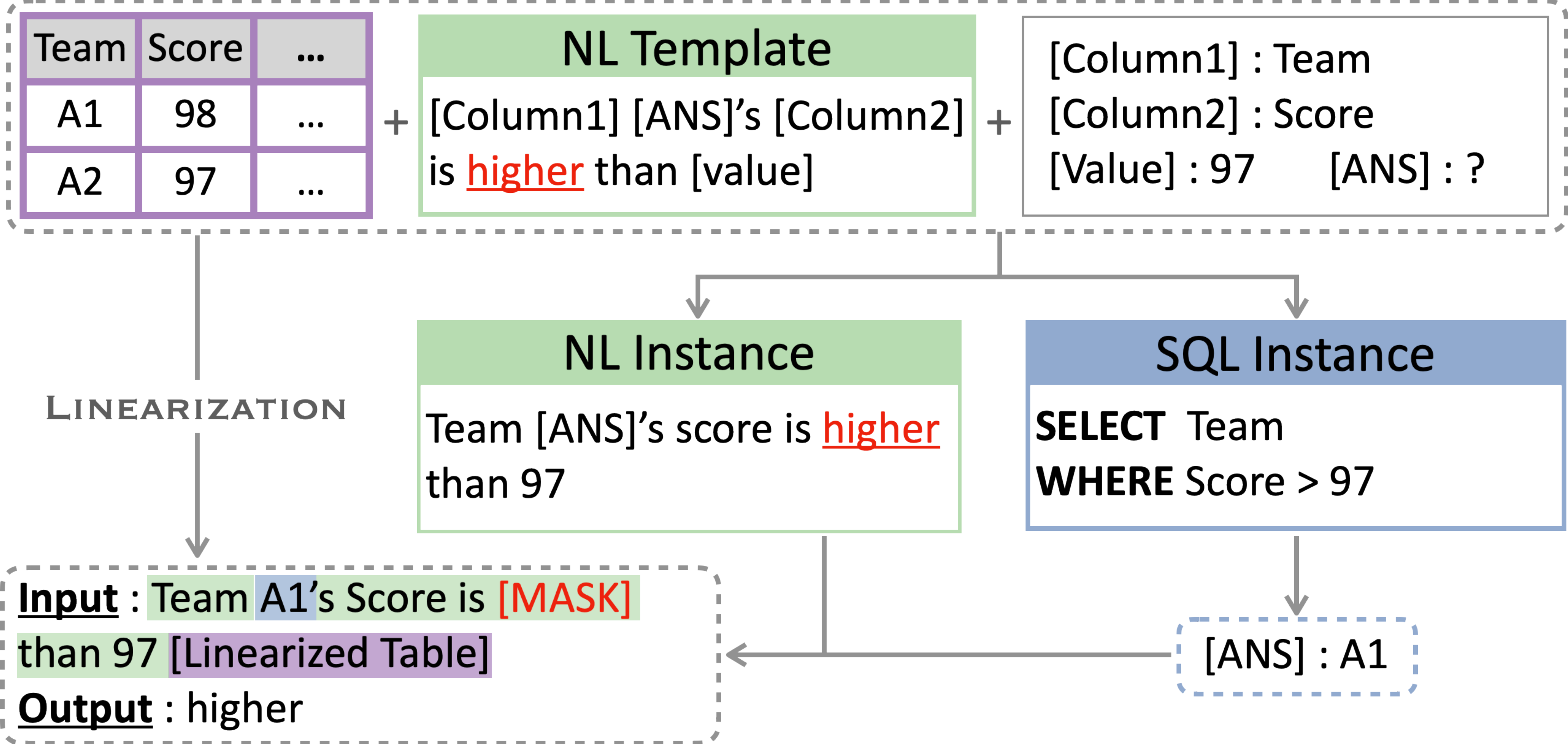}
	\caption{Automatic pre-training corpus generation.}
	\label{fig:data_pipeline}
\end{figure}

\stitle{Sentence Generation.} Figure~\ref{fig:data_pipeline} shows the pipeline of our automatic sentence generation method. 
To ensure that each sentence contains an operation and the operation-aware tokens can be clearly identified, we design \term{NL~Templates} for each table-aware operation type (\eg the \term{NL~Template} in Figure~\ref{fig:data_pipeline} is designed for \term{Comparative}. See more details of the manually designed templates in Appendix~\ref{appendix:Details_of_Pre-training_Corpus}). Each \term{NL~Template} is pre-defined with the position of the operation-aware tokens (\eg \textit{``\underline{higher}''}). 
Unlike fact verification, the sentence in the cloze task must be a correct description of the table. To achieve this, we design an \term{SQL~Template} for each \term{NL~Template}, and these two templates will be instantiated based on the table at the same time. During instantiation, the \term{[Column]} in both templates is replaced by a column header (\eg \term{[Column1]} is replaced by \term{team}) and the \term{[Value]} is instantiated by a cell (\eg \term{[Value]} is replaced by \term{97}). Then, the \term{SQL~Instance} will be automatically executed on the table, and the execution result \term{[ANS]} will be filled in the \term{NL~Instance} to ensure its correctness.

Based on the above method, we generate up to 100 related sentences for each table, depending on the size of the table.
Statistics about the pre-training corpus are given in Table~\ref{tab:cloze_statistic}. We can see that the proportion of each type is different; it mainly depends on how many different expressions a type contains. Taking \term{Aggregation} with the highest proportion as an example: in addition to \textit{``the average of''} in Figure~\ref{fig:overview}, there may also be \textit{``the sum of''}, \textit{``the total amount of''}, etc. 

\stitle{Sentence Polishing.} We notice that using fixed templates for each table could generate unnatural sentences. For example, in Figure~\ref{fig:data_pipeline}, if \term{[Column2]} is not populated with \textit{``score''} but by \textit{``age''}, then the operation-aware token should use \textit{``older''} instead of \textit{``higher''}. Therefore, we introduce a probing-based method to improve the fluency of sentences, which leverages the rich knowledge learned by the LMs implicitly during pre-training. Our main idea is that since BERT-like LMs are pre-trained on extensive textual corpora, their predictions can approximate the natural language expressions used in real-world scenarios. 

Specifically, we identify the context sensitive word $w'$ in each template (\eg \textit{``higher''}), and define a set of candidate values (\eg {\textit{``higher''}, \textit{``more''}, …, \textit{``older''}}) for $w'$. Then, we replace the $w'$ with \term{[MASK]} and leverage a fixed LM to determine the appropriate value for \term{[MASK]} based on context. For example, if \term{[Column2]} is populated as \textit{``age''}, the pre-trained LM calculates probabilities for all candidate values and then selects the one with the highest probability, \eg \textit{``older''}. Please refer to Appendix~\ref{appendix:Details_of_Pre-training_Corpus} for the detailed definition of context sensitive words and their candidate sets.

\begin{table}[!t]
\centering
\begin{tabular}{l|c|c}
\hline
\textbf{Type} & \textbf{\# Sentence-Table} & \textbf{Len (Ans)} \\
\hline
\term{Filter} & 77,609 (6\%) & 3.2 \\
\term{Superlative} & 349,241 (27\%) & 2.6\\
\term{Aggregation} & 388,046 (30\%) & 1.3 \\
\term{Comparative} & 349,241 (27\%) & 1.0 \\
\term{Ordinal} & 103,479 (8\%) & 2.3 \\
\term{Unique} & 25,872 (2\%) & 1.0 \\
\hline
Total & 1,293,488 & 1.8 \\
\hline
\end{tabular}
\caption{\label{tab:cloze_statistic}Statistics of our pre-training corpus, where ``Len (Ans)'' represents the average length of the answer.}
\end{table}


\subsection{Fine-tuning with Select-then-Rank}
\label{sec:finetune}

For fine-tuning, we pre-process the table based on the following two considerations: (i) As mentioned in \citealp{tapasfv}, the sentence-table pairs in the downstream datasets may be too long for pre-trained LMs, and (ii) considering that the disentangled attention mechanism in DeBERTa makes the model more sensitive to the positional information of the input, we assume that it would be easier for the model to capture the sentence-table relationship by putting the most relevant cells in the table closer to the sentence. To address the above two problems, we propose a \textbf{select-then-rank} method to reconstruct the table content.

\stitle{Column-wise Selection.} To make the table size to meet the input length limit of DeBERTaV3, we follow previous work~\cite{tabfact} to only select columns in the table containing entities linked to the statement, which results in a pre-processed table $\Tilde{T}$. Note that the reason for not selecting by rows is that some operations may involve an entire column of cells, \eg \term{Aggregation}.

\stitle{Row-wise Ranking.} To make the sentence and its relevant cells in the table have closer positions, we reorder the table $\Tilde{T}$ by row. Specifically, we slice $\Tilde{T}$ into a set of rows $\{r_1, \ldots, r_m\}$, and rank these rows by their relevance scores $\{p_i\}$, as defined below. Let $\hat{r_i}$ and $\hat{s}$ denote the token sets of row $r_i$ and statement $s$ respectively. The relevance score $p_i$ is given by $|\hat{r_i}\cap \hat{s}|$. Note that we remove the stop-words (\eg \emph{the}) in $\hat{r_i}$ and $\hat{s}$. Finally, we reconstruct the table by ordering the rows in descending order of the relevant scores, before applying 
the table linearization introduced in Section~\ref{sec:deberta}.

\section{Experimental Setup} \label{sec:exp-setup}

\subsection{Datasets}
By using the method introduced in~\ref{sec:pretrain_corpus}, we synthesize 1.2 million sentence-table cloze questions as the pre-training corpus. Specifically, our pre-training corpus contains 20K well-structured tables selected from WikiTables, and sentences are automatically generated from the tables. 

During fine-tuning, we evaluate our model on two widely-adopted table-based fact verification benchmark datasets TabFact~\cite{tabfact} and SEM-TAB-FACTS~\cite{semtabfact}.
\textbf{TabFact} contains 16K tables collected from WikiTables and 118K human-annotated natural language statements, where each statement-table pair is labeled as either \emph{entailed} or \emph{refuted}. TabFact contains statements with two difficulty levels: (i) simple statements corresponding to single rows, and (ii) complex statements involving  multiple rows with table-based operations like \term{Aggregation}. 
\textbf{SEM-TAB-FACTS} contains 2K tables and 4K human-annotated natural language statements. Different from TabFact, these tables are collected from scientific articles in a variety of domains. 
%
We use the official splits in the two benchmarks for evaluation: the training, validation and test sets of TabFact respectively contain 92283, 12792 and 12779 sentence-table pairs; the training, validation and test sets of SEM-TAB-FACTS respectively contain 4506, 423 and 522 sentence-table pairs. In addition, TabFact also holds out a small test set with 2K sentence-table pairs with \emph{human performance}. 



\subsection{Baselines}
We evaluate \pasta with the following ten state-of-the-art methods for table-based fact verification. 

\sstab
{\bf Table-BERT}~\cite{tabfact} adopts templates to linearize a table into an NL sentence, 
and then directly leverages a BERT model to encode the linearized table and the statement.
%

\sstab
{\bf LogicFactChecker}~\cite{logicfactchecker} leverages a sequence-to-action semantic parser to generate a ``program'', \ie a tree with multiple operations, 
%
%
and uses a graph neural network to encode statements, tables, and the generated programs.
%


\sstab
{\bf SAT}~\cite{sat} creates a structure-aware mask matrix to encode the structural data. In particular, it considers recovering the alignment information of tabular data by masking signals of unimportant cells during self-attention.
%

\sstab
{\bf ProgVGAT}~\cite{progvgat} integrates programs and execution into a natural language inference model. This method uses a verbalization with a program execution model to accumulate evidences and constructs a graph attention network to combine various evidences.
%


\sstab
{\bf Tapas}~\cite{tapas} extends BERT with additional structure-aware positional embeddings to represent the tables. \citealp{tapasfv} further pre-train Tapas on counterfactually-augmented and grammar-based synthetic statements.
%

\sstab
{\bf \citealp{joint}} study table-based fact verification in an open-domain setting, and combine a TF-IDF retrieval model with a RoBERTa-based joint reranking-and-verification model.

\sstab
{\bf Tapex}~\cite{tapex} guides the pre-trained BART model to mimic an \term{SQL} executor via an execution-centric table pre-training approach. The pre-training corpus of Tapex is synthesized via sampling \term{SQL} queries from the SQUALL dataset~\cite{squall}.

\sstab
{\bf SaMoE}~\cite{SaMoE} develops a mixture-of-experts network based on the RoBERTa-large model~\cite{roberta}. The MoE network consists of different experts, and then a management module decides the contribution of each expert network to the verification result. 

\sstab
{\bf Volta}~\cite{volta} analyzes how transfer learning and standardizing tables to contain a single header row can boost the effectiveness of table-based fact verification.

\sstab
{\bf LKA}~\cite{lka} studies the sentence-table's evidence correlation. It develops a dual-view alignment module based on the statement and table views to identify the most important words through various interactions.

\subsection{Implementation Details}
Our model is implemented based on the transformer architecture~\cite{transformers}. Specifically, we start pre-training with the public DebertaV3-Large checkpoint\footnote{\url{https://huggingface.co/microsoft/deberta-v3-large}} and optimize the learning objective with Adam~\cite{adam}. Our pre-training process runs up to 400K steps with a batch size of 16 and a learning rate of $1\times10^{-6}$. The complete pre-training procedure takes about 3 days on 2 RTX A6000 GPUs. For fine-tuning, the model runs up to 300K steps with a batch size of 8 and a learning rate of $5\times10^{-6}$.

\begin{table*}[t!]
\centering
\resizebox{\textwidth}{!}{%
\begin{tabular}{llllll}
\hline
\textbf{Model} & \textbf{Val} & \textbf{Test} & \textbf{Simple Test} & \textbf{Complex Test} & \textbf{Small Test}\\
\hline
Table-BERT~\cite{tabfact} & 66.1 & 65.1 & 79.1 & 58.2 & 68.1 \\
LogicFactChecker~\cite{logicfactchecker} & 71.8 & 71.7 & 85.4 & 65.1 & 74.3 \\
SAT~\cite{sat} & 73.3 & 73.2 & 85.5 & 67.2 & - \\
ProgVGAT~\cite{progvgat} & 74.9 & 74.4 & 88.3 & 67.6 & 76.2 \\
\citealp{joint} (Oracle retrieval) & 78.2 & 77.6 & 88.9 & 72.1 & 79.4 \\
Tapas~\cite{tapasfv} & 81.0 & 81.0 & 92.3 & 75.6 & 83.9 \\
Tapex~\cite{tapex} & 84.6 & 84.2 & 93.9 & 79.6 & 85.9 \\
SaMoE~\cite{SaMoE} & 84.2 & 85.1 & 93.6 & 80.9 & 86.7 \\
\hline
DeBERTaV3 & $86.1_{\pm 0.2}$ & $86.2_{\pm 0.1}$ & $92.8_{\pm 0.2}$ & $82.9_{ \pm 0.1}$ & $86.5_{ \pm 0.3}$ \\
\rowcolor[gray]{.9}
\pasta & $\textbf{89.2}_{\pm 0.4}$ & $\textbf{89.3}_{\pm 0.3}$ & $\textbf{96.7}_{\pm 0.2}$ & $\textbf{85.6}_{\pm 0.3}$ & $\textbf{90.6}_{\pm 0.2}$ \\
\hline
Human Performance & - & - & - & - & 92.1 \\
\hline
\end{tabular}
}
\caption{\label{tab:overall_results}Performance on TabFact in terms of binary classification accuracy (\%). The human performance on a small set is from~\citealp{tabfact}. The notation ``-'' indicates that the corresponding values are not listed in the original paper. In addition, models are evaluated with 5 random runs.
}
\end{table*}
\section{Experiments and Results}
\label{sec:results}

\begin{table}[t!]
	\centering
	\resizebox{!}{!}{%
		\begin{tabular}{lll}
			\hline
			\textbf{Model} & \textbf{Val} & \textbf{Test} \\
			\hline
			Volta~\cite{volta} & 74.35 & 73.87 \\
			Tapas~\cite{tapas_semtabfact} & 78.33 & 75.33 \\
			Tapex & 77.53 & 75.47 \\
			LKA~\cite{lka} & 80.34 & 78.54 \\
			\hline
			DeBERTaV3 & 81.85 & 78.92 \\
			\rowcolor[gray]{.9}
			\pasta & \textbf{84.23} & \textbf{84.10} \\
			\hline
		\end{tabular}
		}
	\caption{\label{tab:semtabfact}Performance on SEM-TAB-FACTS in terms of micro-F1 (\%). The experimental results of Tapex is from \citealp{lka}.}
\end{table}

\subsection{Overall Performance}
\label{sec:overall_performance}
Table~\ref{tab:overall_results} summarizes the overall experimental results for various fact verification methods on TabFact. We can see that \pasta achieves the \textbf{new SOTA results} on all splits of TabFact. In particular, on the complex set containing multiple operations, \pasta largely outperforms the previous state-of-the-art by \textbf{4.7} points (\textbf{85.6\% vs. 80.9\%}).

Table~\ref{tab:overall_results} also reports the good performance of DeBERTaV3 on the table-based fact verification task. This result is analogous to the observation in \citealp{joint} that RoBERTa can yield strong performance exceeding the previous closed-setting (77.6\% vs. 74.4\%). Both results illustrate that the BERT-like model pre-trained on textual data can also perform well on linearized tabular data. 
However, \pasta surpasses DeBERTaV3 by 3.1 points on the test set (89.3\% vs. 86.2\%), \ie 3.9 points and 2.7 points on the simple test set and complex test set respectively. The experimental result shows that \pasta endows DeBERTaV3 with more powerful statement-table reasoning ability, which is very crucial for fact verification.

Table~\ref{tab:semtabfact} shows that \pasta outperforms all baseline models by large margins on the SEM-TAB-FACTS dataset. In particular, \pasta significantly surpasses the DeBERTaV3 model by 5.2 points (84.1\% vs 78.9\%) on the test set. This shows that although our pre-training corpus only contains tables from Wikipedia, it can be applied to other domains, such as tables from scientific articles included in the SEM-TAB-FACTS dataset.

\subsection{Impact of Operation Aware Pre-training}
\label{sec:impact_of_opearion_aware_pretraining}
\begin{figure}[t!]
	\centering
	\includegraphics[width=1.\columnwidth]{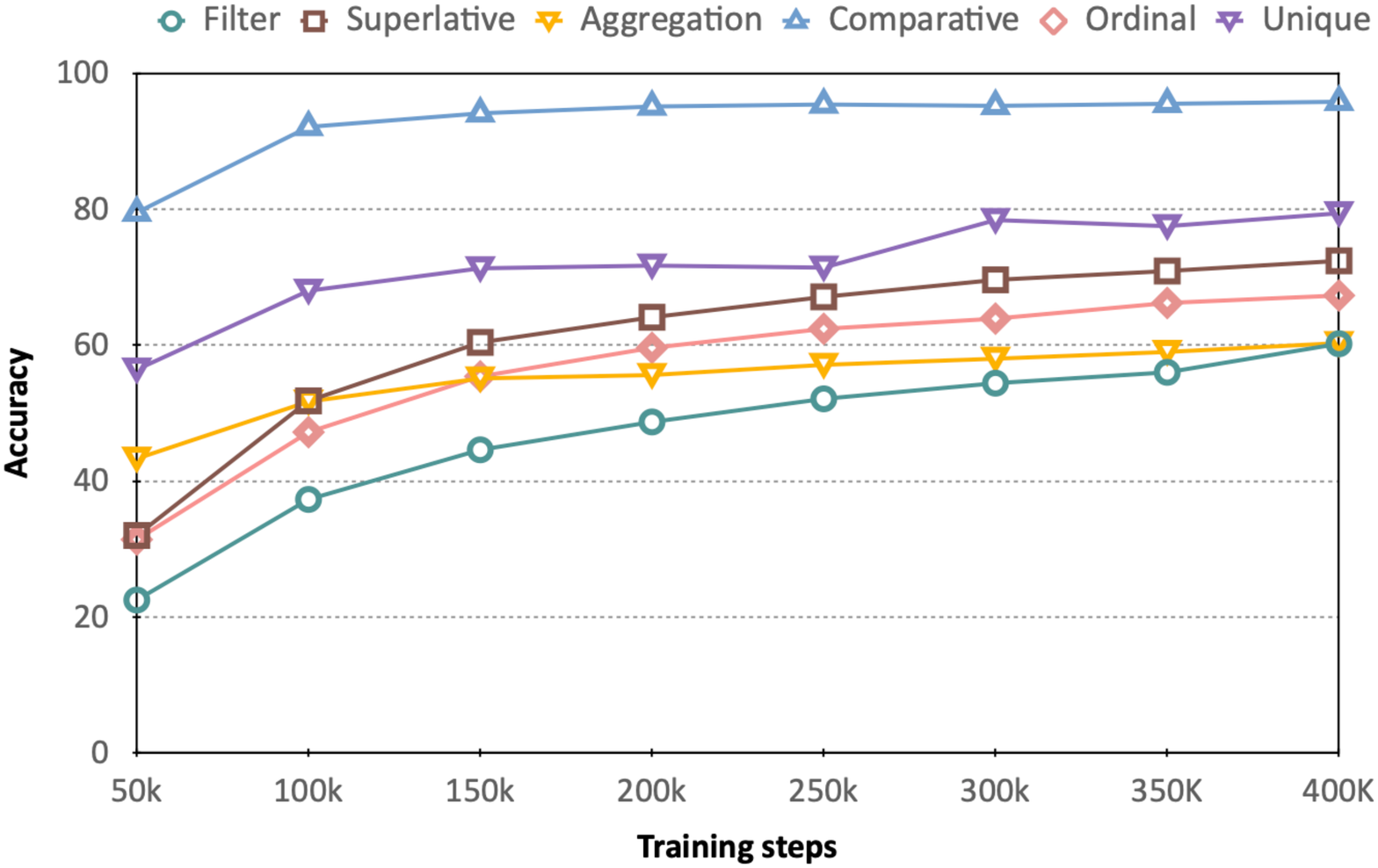}
	\caption{Accuracy on operation-aware cloze tasks at different training steps. For each operation, the size of its test set is 1K where the test set does not contain any tables from the training set.}
	\label{fig:cloze_performance}
\end{figure}

\stitle{Performance on cloze pre-training.}
As described in Section~\ref{sec: pretrain_task}, \pasta is pre-trained on table-operations aware cloze tasks to be capable of reasoning about {table-aware operations} over tables.
%
%
%
To explore whether the model has learned such ability, we generate six test sets corresponding to different operation types, and evaluate the performance of \pasta in various steps during pre-training.
The experimental results are reported in Figure~\ref{fig:cloze_performance}.
Overall, after 400K steps, \pasta can correctly complete more than $60\%$ sentence-table cloze questions with various types.
More specifically, with increasing the steps, \pasta is firstly capable of reasoning about \term{Comparative} operations, and finally mastering \term{Aggregation} and \term{Filter} operations.
This may be attributed to the difficulty of the operations (\eg \term{Aggregation} is harder than other types) and the length of token span that needs to be predicted (As shown in Table~\ref{tab:cloze_statistic}, \term{Filter} needs to predict more tokens than other types).



\begin{table*}[!t]
\centering
\resizebox{\textwidth}{!}{%
\begin{tabular}{c|c|c|>{\columncolor[gray]{.9}}c}
\hline
\textbf{Operation} & \textbf{Example in TabFact} & \textbf{DeBERTaV3} & \textbf{\pasta} \\
\hline
\term{Filter} & the blues and penguins game on march 20 , score was 2 - 4 & 88.0 & \textbf{90.7} (+2.7) \\
\term{Superlative} & pacific national has the highest number in class & 85.9 & \textbf{86.7} (+0.8) \\
\term{Aggregation} & the average amount of points among all teams is 29 & 81.0 & \textbf{84.5} (+3.5) \\
\term{Comparative} & ian woosnam placed higher than craig parry & 85.2 & \textbf{86.2} (+1.0) \\
\term{Ordinal} & the second largest number of runs was 8529 & 83.8 & \textbf{86.9} (+3.1) \\
\term{Unique} & there are 5 different nations in the tournament & 74.2 & \textbf{79.1} (+4.9) \\
\hline
\end{tabular}
} %
\caption{\label{tab:op_test}Binary classification accuracy (\%) 
on sentence-table pairs containing different types of operations. The six sets are sampled from TabFact based on trigger words, and each set contains 200 sentence-table pairs. 
}
\end{table*}

\stitle{Operation understanding on fact verification.} We further analyze whether the model can utilize the reasoning ability learned from pre-training for our downstream task, \ie table-based fact verification.
To this end, we evaluate \pasta on test sets of different operation types. We split the test set of TabFact according to the \emph{trigger words} in the statement, which are defined in Appendix~\ref{appendix:Trigger_Words_Definition}, \eg ``\textit{highest}'' and ``\textit{lowest}'' related to the \term{Superlative} type.
%
We control the size of each test set as 200, while ensuring that these sets have no overlap.
We compare \pasta and DeBERTaV3 on these test sets, and the results are shown in the Table~\ref{tab:op_test}. We can see that \pasta outperforms DeBERTaV3 on every test set, especially on the \term{Aggregation} type.
%
Note that we did not use any fine-tuning strategy on both models, 
and thus all the improvements of \pasta are to be attributed to our table-operations aware pre-training strategy.

\begin{table}
	\centering
	\resizebox{\columnwidth}{!}{%
		\begin{tabular}{l|llll}
			\hline
			\textbf{Scheme} & \textbf{Val} & \textbf{Test} & \textbf{Simple} & \textbf{Complex} \\
			\hline
			MLM  & $84.8_{\pm 0.2}$ & $84.9_{\pm 0.2}$ & $92.5_{\pm 0.1}$ & $81.2_{\pm 0.2}$ \\
			\rowcolor[gray]{.9}
			\pasta & $\textbf{87.0}_{\pm 0.1}$ & $\textbf{87.9}_{\pm 0.2}$ & $\textbf{94.0}_{\pm 0.2}$ & $\textbf{84.9}_{\pm 0.2}$ \\
			\hline
		\end{tabular}
		}
	\caption{\label{tab:MLM}Ablation study for the masking scheme. MLM uses random masking and \pasta uses table-operations aware masking. To avoid co-effects, we didn't use any data pre-processing method on MLM or \pasta. Models are evaluated with 5 random runs. }
\end{table}

\stitle{Comparison with Masked Language Modeling.}
%
%
We compare \pasta with the random masking scheme in Masked Language Modeling (MLM).
%
For MLM, we randomly mask 15\% of the tokens in a sentence-table pair, of which 10\% of the masked tokens remain unchanged, 10\% are replaced with randomly picked tokens, and the remainders are replaced with the [${\tt {MASK}}$] token. For \pasta, we use the masking strategy introduced in Section~\ref{sec: pretrain_task}, which only masks the operation-aware span in the sentence. We pre-train both MLM and \pasta on DeBERTaV3.
%
Considering pre-training efficiency, for both MLM and \pasta, we set the training step as 140K.
Table~\ref{tab:MLM} shows the fine-tuned results of MLM and \pasta on the TabFact dataset. We can see that \pasta outperforms MLM by a large margin on the complex set. This improvement further proves that our table-operations aware pre-training task helps the high-order symbolic reasoning in the complex set. 
By also observing Table~\ref{tab:overall_results}, we find that the impact of MLM decreases slightly on the basis of DeBERTaV3 (84.9\% vs. 86.2\%). This may be because
the random masking scheme in MLM does not work for sentence-table joint understanding, as we have already analyzed in Section~\ref{sec: pretrain_task}.

\subsection{Impact of Select-then-Rank}
To verify the effectiveness of the select-then-rank method, we conduct experiments with column-wise selection and row-wise ranking on the TabFact dataset. The results of the experiment are shown in Table~\ref{tab:select-then-rank}. We can see that the row-wise ranking strategy is more effective than the column-wise selection strategy. The main reason may be that the disentangled attention mechanism in DeBERTa makes the model more sensitive to the positional information of the input. The row-wise ranking strategy can put the most relevant cells in the table closer to the sentence, and thus the model can more effectively capture the sentence-table relationship.

\begin{table}[t!]
	\centering
	\resizebox{!}{!}{%
		\begin{tabular}{lcccc}
			\hline
			\textbf{Method} & \textbf{Val} & \textbf{Test} & \textbf{Simple} & \textbf{Complex} \\
			\hline
			\pasta  & \textbf{$89.2$} & \textbf{$89.3$} & \textbf{$96.7$} & \textbf{$85.6$} \\
			w/o col & $88.9$ & $89.0$ & $95.8$ & $85.5$ \\
			w/o row & $88.2$ & $88.4$ & $95.1$ & $84.7$ \\
			\hline
		\end{tabular}
		}
	\caption{\label{tab:select-then-rank}Ablation study for the select-then-rank strategy. ``w/o col'' means that the column-wise selection strategy is not used, and ``w/o row'' means that the row-wise ranking strategy is not used.}
\end{table}

\subsection{Error Analysis}
To analyze the errors of \pasta for table-based fact verification, we analyze the sentence-table pairs that \pasta predicts incorrectly in the TabFact dataset. Specifically, we consider the size of the tables and the complexity of the operations in the statements. Table~\ref{tab:error} presents some basic statistics about the subset of the test where \pasta makes mistakes.
For comparison, we also list the basic statistics about DeBERTaV3's error set and the full test set. 
We have the following observations.
%
(1) Our models may not perform well on large tables.
%
Concretely, although \pasta reduces the impact of large tables on the fact verification task compared to DeBERTaV3 (97.5 vs. 107.4), the impact of large tables on \pasta still exists compared to the average table size in the test set (97.5 vs. 89.0). 
(2) The number of operations in the statement is also an important cause of errors: the proportion of statements with multiple operations in \pasta's error set (16.5\%) is larger than that of the overall test set (11.3\%). Thus, \pasta correctly verifies most of the statements that contain only a single operation, but it still encounters difficulty to verify statements that contain multiple types of operations.

\begin{table}[t!]
\centering
\resizebox{\columnwidth}{!}{%
\begin{tabular}{lcccc}
\hline
 & \textbf{\# Row} & \textbf{\# Col} & \textbf{\# Cell} & \textbf{\% Mul-Ops} \\
\hline
All Test & 6.2 & 14.3 & 89.0 & 11.3 \\
DeBERTaV3 & 6.4 & 16.8 & 107.4 & 13.4\\
\rowcolor[gray]{.9}
\pasta & 6.4 & 15.5 & 97.5 & 16.5\\
\hline
\end{tabular}
}
\caption{\label{tab:error} Statistics of data collected from \pasta and DeBERTaV3's error sets. Proportion of statements that contain more than two operations is marked as ``\% Mul-Ops''.
``\# Row'', ``\# Col'', ``\# Cell'' respectively represent the average numbers of rows, columns, and cells.}
\end{table}

\section{Related Work}

\paragraph{Sentence-Table Joint Understanding}

Many 
tasks, such as table question answering, table search, table-to-text, and table-based fact verification, require understanding tables and an \term{NL} sentence jointly.
TAPAS~\cite{tapas} and FORTAP~\cite{fortap} design sentence-table joint pre-training tasks. TAPAS further leverages Whole Word Masking (WWM) to learn better representations of tables, while FORTAP leverages Numerical Reference Prediction (NRP) and Numerical Calculation Prediction (NCP) on a large corpus of spreadsheet formulas.
These methods are closest to ours, but one focuses on the contextual information of \term{NL} and the other on the statistical characteristics of tables. Our method considers both, and improves the joint understanding of the model by means of operation-aware cloze tasks.

\paragraph{(Table-based) Fact Verification}

Program-driven methods such as LogicFactChecker~\cite{logicfactchecker} and ProgVGAT~\cite{progvgat} mainly focus on explicitly capturing logical operations in statements and representing them using a graph neural network to support fact verification. PLM-driven methods such as Table-BERT~\cite{tabfact}, TAPAS~\cite{tapasfv}, and SaMoE~\cite{SaMoE} perform table-based fact verification tasks as an NLI task and apply a BERT-like model to encode sentence-table pairs. Our approach is more similar to SaMoE~\cite{SaMoE}. 
The main difference is that SaMoE uses different experts to solve different types of operations, while our method assumes that the operation combinations in statements are complex and diverse. We directly inject the atomic operations into the model through operation-aware pre-training, and then fine-tune the model to learn the various operation combinations on downstream datasets. However, \pasta is compatible with the mixture-of-experts framework, and thus we would evaluate PASTA with MoE structure on table-based fact verification datasets in the future.

\section{Conclusion and Future Work}

We introduced \pasta, a table-operations aware pre-training approach to train LMs for better performing fact verification over tables.
\pasta achieved new SOTA results on two widely-adopted table-based fact verification benchmark datasets, TabFact and SEM-TAB-FACTS.
Future work should explore how to address the challenges of more complex operations and large tables in fact verification.

\section*{Limitations}
The first limitation of our work is that our synthetic pre-training corpus may lack diversity. As explained in Section~\ref{sec:pretrain_corpus}, to ensure the correctness and controllability of these sentences, we generate the pre-training corpus using human-designed natural language templates. While our insight is that only atomic operations need to be learned at pre-training, which reduces the need for diversity, generating high-quality sentences with both diversity and controllability to support self-supervised learning is still a direction worth exploring. 

The second limitation of our work is that fact verification is only supported on a single table. Although the TabFact~\cite{tabfact} dataset we used assumes that each statement can be verified by a single table, a more realistic scenario would be to combine information from multiple tables. Exploring how to do this effectively and to overcome the limitation of the input length of BERT-like models is a important direction for future work.

\section*{Ethics Statement}
\stitle{Dataset Collection} For the pre-training dataset, we use the publicly available WikiTables~\cite{wikitables} dataset as the table source and select high-quality relational tables from it. Then we use these tables to generate entailed statements, instead of collecting statements from the web. For the fine-tuning dataset, we use the publicly available datasets, TabFact~\cite{tabfact} and SEM-TAB-FACTS~\cite{semtabfact}. 

\stitle{Intended Use and Misuse Potential} The goal of the fact verification task is to help identify misinformation. Our work focuses on verifying the statements based on analysis over tables and aims to pre-train language models to be aware of common table operations, such as {aggregation} over a column or comparing two tuples. It should be noted that while we treat the tables from TabFact and SEM-TAB-FACTS as trustworthy sources of evidence in the experiments, we do not assume that all of the tables in the network are trustworthy and unbiased. So, this work could also be misused through fact verification on unreliable or socially biased tables. 

\stitle{Broader Impact} First, our operation-aware pretraining approach potentially has a broad impact on sentence-table joint understanding tasks. Second, we also notice that training a large-scale pre-trained language model requires the use of GPUs/TPUs for training, which contributes to global warming. However, we're pre-training from a public checkpoint, not from scratch. And our pre-training dataset contains only 1.2 million examples, which is very small compared to other related work~\cite{tapasfv,tapex}. 
\section*{Acknowledgements}
This work was partly supported by the NSF of China (62122090, 62072461, and U1911203) and the fund for building world-class universities (disciplines) of Renmin University of China.
\bibliography{refs/TaRA}
\bibliographystyle{acl_natbib}
\appendix
\begin{table*}
\centering
\resizebox{\linewidth}{!}{
\begin{tabular}{|c|m{9cm}|m{9cm}|}
\hline
\textbf{Operation} & \textbf{NL Template} & \textbf{SQL Template} \\
\hline
\multirow{2}{*}{\texttt{Filter}} & [Value2]'s [Column1] is \underline{[ANS]}. & SELECT [Column1] FROM T WHERE [Column2] = [Value1] \\
\cline{2-3}\xrowht{20pt}
 & the [Column1] of [Value2] is \underline{[ANS]}. & SELECT [Column1] FROM T WHERE [Column2] = [Value1] \\
\hline\xrowht{20pt}
\multirow{2}{*}{\texttt{Superlative}} & the \textcolor{red}{highest} [Column1] is \underline{[ANS]} & SELECT MAX([Column1]) FROM T \\
\cline{2-3}
 & \underline{[ANS]} has the \textcolor{red}{highest} [Column2] of all [Column1] & SELECT [Column1] FROM T ORDER BY [Column2] DESC LIMIT 1 \\
\hline
\multirow{2}{*}{\texttt{Aggregation}} & the sum of [Column1] when [Column2] is [Value2] is \underline{[ANS]}. & SELECT SUM([Column1]) FROM T WHERE [Column2] = [Value2] \\
\cline{2-3}
 & the average of [Column1] when [Column2] is [Value2] is \underline{[ANS]}. & SELECT AVG([Column1]) FROM T WHERE [Column2] = [Value2] \\
\hline
\multirow{2}{*}{\texttt{Comparative}} & [Column1] [ANS]'s [Column2] is \underline{\textcolor{red}{higher}} than [Value2] & SELECT [Column1] FROM t WHERE [Column2] > [Value2] \\
\cline{2-3}
 & [ANS] has \underline{\textcolor{red}{higher}} [Column2] than [Value2] & SELECT [Column1] FROM t WHERE [Column2] > [Value2] \\
\hline
\multirow{2}{*}{\texttt{Ordinal}} & \underline{[ANS]} has the second \textcolor{red}{highest} [Column2] & SELECT [Column1] FROM T WHERE [Column2] < ( SELECT MAX([Column2]) FROM T ) ORDER BY [Column2] DESC LIMIT 1 \\
\cline{2-3}
 & \underline{[ANS]} has the second \textcolor{red}{lowest} [Column2] & SELECT [Column1] FROM T WHERE [Column2] > ( SELECT MIN([Column2]) FROM T ) ORDER BY [Column2] ASC LIMIT 1 \\
\hline\xrowht{20pt}
\multirow{2}{*}{\texttt{Unique}} & there are \underline{[ANS]} different [Column1] on the list. & SELECT COUNT( DISTINCT [Column1] ) FROM T \\
\cline{2-3}\xrowht{20pt}
 & the total number of different [Column1] is \underline{[ANS]}. & SELECT COUNT( DISTINCT [Column1] ) FROM T \\
\hline
\end{tabular}
}
\caption{\label{tab:cloze_example}~Examples of NL Templates and SQL Templates for each table-aware operation type. The words in each NL Template that need to be polished by the pre-trained LM are marked in \textcolor{red}{red}. The operation-aware tokens that need to be predicted during pre-training are \underline{underlined}.}
\end{table*}

\begin{table*}
\centering
\resizebox{!}{!}{
\begin{tabular}{|c|p{13.5cm}|}
\hline
$w'$ & Candidate Set \\
\hline
``highest'' & {``highest'', ``most'', ``biggest'', ``largest'', ``oldest'', ``greatest'', ``heaviest'', ``longest'', ``tallest''} \\
\hline
``lowest'' & {``lowest'', ``least'', ``smallest'', ``youngest'', ``shortest''} \\
\hline
``higher'' & {``higher'', ``more'', ``bigger'', ``larger'', ``older''} \\
\hline
``less'' & {``less'', ``smaller'', ``lower'', ``younger''} \\
\hline
\end{tabular}
}
\caption{\label{tab:context-sensitive-words}~Four context-sensitive word candidate sets. $w'$ refers to the context-sensitive word in the NL templates.}
\end{table*}

\begin{table*}
\centering
\resizebox{!}{!}{
\begin{tabular}{|m{2cm}|p{13cm}|}
\hline
Operation & Trigger Words \\
\hline
\texttt{Filter} & {``is'', ``was'', ``are'', ``were''} \\
\hline
\texttt{Superlative} & {``lowest'', ``least'', ``smallest'', ``youngest'', ``shortest'', ``first'', ``best'', ``newest'', ``latest''} \\
\hline
\texttt{Aggregation} & {``average'', ``count'', ``sum'', ``total''} \\
\hline
\texttt{Comparative} & {``than'', ``higher'', ``more'', ``bigger'', ``larger'', ``older'', ``less'', ``smaller'', ``lower'', ``younger'', ``same'', ``equal''} \\
\hline
\texttt{Ordinal} & {``second'', ``third'', ``fourth''} \\
\hline
\texttt{Unique} & {``different''} \\
\hline
\end{tabular}
}
\caption{\label{tab:trigger-words}~Trigger words for each table-aware operation type.}
\end{table*}

\section{Details of Pre-training Corpus}
\label{appendix:Details_of_Pre-training_Corpus}
For the six table-aware operation types, we design a total of 50 NL-SQL template pairs to generate sentence-table instances for pre-training. Table~\ref{tab:cloze_example} shows two examples of each type of operations. We first populate the NL-SQL template pairs with the specific content in the table $T$, and then the context sensitive words $w'$ in the NL template are selected by a fixed pre-trained LM in its synonym set to improve the fluency of the generated sentences. We define four context-sensitive word candidate sets, which are presented in Table~\ref{tab:context-sensitive-words}.

\section{Trigger Words Definition}
\label{appendix:Trigger_Words_Definition}

Table~\ref{tab:trigger-words} shows the trigger words we have used to identify different operation types in Section~\ref{sec:impact_of_opearion_aware_pretraining}. These trigger words are expanded from the trigger words defined in TabFact~\cite{tabert}, and are then classified according to the six table-aware operation types introduced in this paper.

\end{document}